\documentclass{article}

\usepackage[preprint]{neurips_2019}

\usepackage[utf8]{inputenc} % allow utf-8 input
\usepackage[T1]{fontenc}    % use 8-bit T1 fonts
\usepackage{hyperref}       % hyperlinks
\usepackage{url}            % simple URL typesetting
\usepackage{booktabs}       % professional-quality tables
\usepackage{amsfonts}       % blackboard math symbols
\usepackage{nicefrac}       % compact symbols for 1/2, etc.
\usepackage{microtype}      % microtypography
\usepackage[ruled,vlined]{algorithm2e}

\usepackage{float}

\usepackage{algpseudocode}% http://ctan.org/pkg/algorithmicx
\usepackage{color}
\usepackage{wrapfig}
\usepackage{tabularx}
\usepackage{makecell}
\usepackage{graphicx}
\usepackage{subcaption}

\usepackage{textcomp}
\UndeclareTextCommand{\textsection}{T1}
\SetProtrusion[context=footnote]{font=TS1/*/m/n/scriptsize}{%
  \textasteriskcentered={,650},
  \textsection={,650},
  \textdagger={,650},
  \textdaggerdbl={,650},
  \textparagraph={,650},
  \textbardbl={,650},
  }

\newcommand{\cut}[1]{}

\usepackage[textsize=footnotesize,color=green!40]{todonotes}

\usepackage{cancel}

\title{Actor Critic with Differentially Private Critic}
\author{Jonathan Lebensold, Borja Balle, William Hamilton, Doina Precup}
\date{August 2019}

\author{%
  Jonathan Lebensold\thanks{jonathan.maloney-lebensold@mail.mcgill.ca} , William Hamilton\thanks{wlh@cs.mcgill.ca}, Borja Balle\thanks{borja.balle@gmail.com}, Doina Precup\thanks{dprecup@cs.mcgill.ca}\\
  \textsuperscript{\textasteriskcentered}\textsuperscript{\textdagger}\textsuperscript{\textsection} McGill University, Mila - Qu\'ebec Artificial Intelligence Institute \\
  \textsuperscript{\textsection}Google DeepMind \\
}

\usepackage{graphicx}

\usepackage{xspace}

\def\p(#1|#2){p(#1\,|\,#2)}
\def\w{{\bf w}}

\def\v{{\bf v}}

\def\={\!=\!}

\begin{document}

\maketitle

\begin{abstract}
  Reinforcement learning algorithms are known to be sample inefficient, and often performance on one task can be substantially improved by leveraging information (e.g., via pre-training) on other related tasks. In this work, we propose a technique to achieve such knowledge transfer in cases where agent trajectories contain sensitive or private information, such as in the healthcare domain. Our approach leverages a differentially private policy evaluation algorithm to initialize an actor-critic model and improve the effectiveness of learning in downstream tasks. We empirically show this technique increases sample efficiency in resource-constrained control problems while preserving the privacy of trajectories collected in an upstream task.
\end{abstract}

\section{Introduction}

 While reinforcement learning (RL) is an attractive framework for modeling decision making under uncertainty, the sample inefficiency challenges are well known \cite{wang2017ACER}, and in some case can be surmounted by, for example, using a simulator \textit{in silico} \cite{fox2019reinforcement} or relying on some form of transfer learning \cite{taylor2009transfer,Parisotto2016}. These solutions, however, rarely account for real-world constraints arising in tasks where data privacy must be addressed. Examples include hospitals sharing patient data to improve clinical decision-making, or navigational information collected from agents in the real-world \cite{fridman2018deeptraffic,zhu2017target}. Thus, as more real-world problems are modelled as use cases for RL algorithms, privacy-preserving knowledge transfer between agent environments will become a deployment requirement.

\begin{figure}[t]
    \center
    \includegraphics[width=6.5cm]{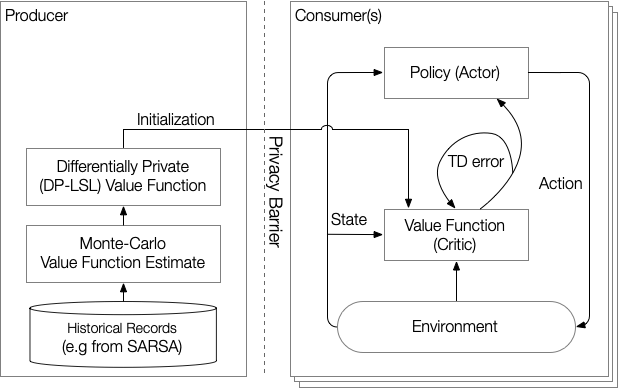}
    \caption{DP Producer/Consumer preserves privacy with a single transfer from the producer.  }
    \label{fig:ac}
\end{figure}

 Differential privacy (DP) is a robust privacy-preserving technique for data analysis algorithms \cite{Apple2017, al2019privacy}. Previous works on DP for sequential decision-making tasks have focused on the (contextual) bandits setting \cite{Shariff2018,tossou2016algorithms, tossou2017achieving, gajane2017corrupt}, or developed RL methods which treat the rewards as sensitive \cite{Wang2019}. We argue that such approaches do not address situations where a trusted aggregator wishes to use historical, potentially sensitive data to bootstrap an RL algorithm to learn on an untrusted environment. For example, aggregating sensitive patient information to train an agent which could then be shared and improved (e.g.\ through personalization) on a smaller, local dataset. In this paper we model such a scenario by assuming a number of untrusted agents (consumers) whose goal to solve an RL task in a sample-efficient manner by leveraging information obtained from a trusted aggregator (the producer). The setup we consider is described in Figure~\ref{fig:ac}.
 
Privacy risks are well studied in the supervised learning setting and DP provides safeguards against an attacker attempting to learn whether an individual record is included in the training set \cite{Carlini2018,Abadi2016}. In the context of RL, previous work has demonstrated how an attacker can infer information about training trajectories from a policy \cite{Pan2019}. The clinical example we use to motivate our work is inspired by recent successes of RL in the context of sepsis treatments \cite{Komorowski2018}.

How best to transfer knowledge from one RL agent to another, even without privacy considerations, remains an area of active research. Informally, any technique where an algorithm is parametrized based on previous training can be considered a form of transfer learning \cite{bengio2007greedy}. The goal of most transfer learning is to use previously learned knowledge to speed up learning in a related task. Multi-task RL \cite{Teh2017} and multi-agent RL \cite{Lowe2017}, model distillation \cite{Rusu2015} and meta-learning \cite{Finn2017} are strategies considered when the task may vary between environments or where a prior is assumed to increase sample efficiency. Transfer learning is also frequently used in supervised learning for vision and speech tasks \cite{oquab2014learning,yosinski2014transferable, Onu2019}.
%A common example occurs in image classification tasks for a specialized class set: a deep neural network's model weights are initialized with a pre-trained model to improve sample efficiency. 

% what's the key idea in our approach?

In this work, we explore transfer learning in RL under DP constraints. 
Specifically, we investigate how actor-critic methods perform when initialized using a privatized First-Visit Monte Carlo estimate \cite{Sutton2018} of the value function. A trusted data aggregator, called the \textit{producer} in our algorithm, uses DP-LSL  \cite{Balle2016}, a differentially private policy evaluation algorithm, to learn a value function $\hat{v}(s,\w)$. Actor-critic, a commonly used policy gradient method in RL, enables our agent, the \textit{consumer}, to iteratively improve an action-value function $Q_{\pi}(s,a, \theta)$ and a state-value function $\hat{v}(s,\w)$. Our method uses the output of DP-LSL to initialize the actor-critic algorithm, effectively transfering knowledge between the producer and the consumers while preserving the privacy of the trajectories collected by the producer. Such an approach is desirable when the environment of the consumer may be considered data-limited.

%We take actions according to a softmax distribution \cite{Sutton2018}.

 %Decision-making systems that must work with uncertainty, whether due to a lack of data or noisy samples, are becoming increasingly prevalent. The Reinforcement Learning setting provides algorithms that are well suited to environments where sequential decision making, learning through action and explainability are required. Our two main contributions are as follows: a well-motivated use case for applying differential privacy with the online Reinforcement Learning setting, and a series of experimental results that measure the utility of leveraging task transfer with Differential Privacy.

%In our algorithm the critic (state-value function) and actor (action-value function) are parameterized as a linear regression of a weight vector $\mathbb{R}^d$ for the $d$-dimensional size of our state-space. The critic is initialized with a copy of the weight vector, $\hat{\theta}^{\lambda}_X$ provided by the producer.
%This statistical difference enables all data contributors to benefit from a certain degree of probable deniability \cite{Dwork2014} as to whether they were in the dataset to begin with. 

\section{Differentially Private Policy Evaluation with DP-LSL}

Differential privacy is achieved by introducing carefully calibrated noise into an algorithm. The goal of a differentially private algorithm is to bound the effect that any individual user's contribution might have on the output while maintaining proximity to the original output. By limiting individual contributions, the potential risk of an adversary learning sensitive information about any one user is limited.

A \textit{user} must be defined in order to properly calibrate noise.  
We assume each user contributes a single trajectory $x_i$ to a database of records $X$ collected by the producer. Each $x_i = ( (s_1, a_1, r_1), ... (s_T, a_T, r_T))_i$ represents a trajectory of states, actions and rewards in $[R_{\min},R_{\max}]$. For example, patient treatment decisions can be expressed as actions selected from a set $\mathcal{A}$ and health readings as observations from a state-space $\mathcal{S}$, while rewards model the treatment outcome. The number of trajectories in $X$ is denoted by $m$.  The set of a states visited by a single trajectory is denoted by $\mathcal{S}_{x_{i}}$. The First-Visit Monte Carlo estimate $F_{X,s}$ of the value of a state $s$ is defined as the empirical average over $X$ of $\gamma$-discounted sums of rewards $F_{x_i,s}$ observed from the first visit to $s$ on each trajectory $x_i$.\footnote{Trajectories where $s \notin \mathcal{S}_{x_i}$ are ignored when computing $F_{X,s}$.}

DP-LSL \cite{Balle2016} is one of the few DP algorithms to support policy evaluation and provide a theoretical privacy and utility guarantee. By treating the estimation of the value function as a regularized least-squares regression problem based on Monte-Carlo estimates, $F_{X, s}$, we can guarantee a limit on the influence of each trajectory to the value function. 
%This algorithm will produce a strongly convex optimization problem. 
DP-LSL achieves differential privacy by adding Gaussian noise to the output of this regression problem; the noise is calibrated in a \emph{data-dependent} manner to achieve $(\epsilon,\delta)$-DP by using the smooth sensitivity framework \cite{Nissim2011}.
% in a region of the space. Be able to take advantage of the inherent properties of the 

More formally, to find the parameter vector $\theta \in \mathbb{R}^m$ representing the value function, DP-LSL minimizes the objective function $J^\lambda_X(\theta)$ below, which includes a ridge penalty with $\lambda > 0$:
\begin{equation}
J^\lambda_{X}(\theta)=\frac{1}{m} \sum_{i=1}^{m} \sum_{s \in \mathcal{S}_{x_{i}}} \rho_{s}\left(F_{x_{i}, s}-\phi_{s}^{\top} \theta\right)^{2} + \frac{\lambda}{2 m}\|\theta\|_{2}^{2}
\label{eq_obj}
\end{equation}.

The regression weights $0 \leq \rho_s \leq 1$ represent any initial prior/importance that we may ascribe to each state, eg.\ depending on how frequently they are visited.
This least-squares problem can be solved in closed-form to find a parameter vector $\theta^\lambda_X \in \mathbb{R}^d$ yielding the value function $\hat{V}^\pi = \Phi \theta^\lambda_X$, where $\Phi$ is a given feature matrix containing the features $\phi_s$ for each state. Gaussian noise is then applied to the result.
The utility analysis in \cite{Balle2016} suggests that the regularization parameter $\lambda$ must be carefully chosen as a function, which depends only on the number of trajectories $m$.

\section{Actor Critic with Differentially Private Critic}

Our proposed algorithm comprises of two phases: a \textit{producer} that uses historical data --- considered confidential --- to evaluate a policy, i.e.\ obtain the associated value function; and a \textit{consumer} that uses an actor-critic algorithm initialized with the value function provided by the producer. Intuitively, such a prior should help the actor make initial estimates of actions taken. While we restrict ourselves to actor-critic, any algorithm that incorporates a value function could be used for the consumer phase.

\begin{algorithm}[H]
\SetAlgoLined
\KwIn{$X, \Phi, \gamma, R_{max}, \rho, \lambda, \epsilon, \delta $}
 
Compute First-Visit Monte-Carlo Estimate $F_{x,s}$
 
Compute DP-LSL: $\hat{\theta}^{\lambda}_X$
 
Let $\w \leftarrow \hat{\theta}^{\lambda}_X$
 
 Run Actor-Critic with Critic $\hat{v}(s,\w)$
 \caption{Actor-critic with Differentially-Private Critic}
\end{algorithm}

The producer is therefore responsible for policy evaluation i.e., attempting to learn the state-value function $V^{\pi}$ for a for a given policy $\pi$. 
Empirical results come from collecting sample trajectories $x_i$ using a learning algorithm (SARSA) \cite{Sutton2018}. While any learning algorithm could be used in practice, SARSA is appealing due to its relative simplicity and convergence properties at the limit.

%Actor-critic with Differentially-Private Critic relies on the following inputs. 

\section{Empirical Setup}

Our empirical results come from 2 domains: an MDP domain (100 states with two actions) and the OpenAI Gym \cite{openAIGym} environment \textit{Taxi-V2} \cite{Dietterich2000}.
These experiments compare the benefit of value function transfer in consumer phase. The producer phase outputs a least-squares approximation of a differentially-private value function. By letting $\w \leftarrow \hat{\theta}^{\lambda}_X$, we can then initialize $ \hat{v} ( s ,\w )$. 
A ridge-regularization parameter $\lambda$ is parameterized based on a regularization term $r$ where $ \lambda = rm^p $, $p=0.5$ and $r=200$. We fix privacy parameter $\delta = 1/m$ and vary $\epsilon$. 
%($\lambda$ increases in proportion to $m$)

\begin{figure}
    \includegraphics[width=\textwidth]{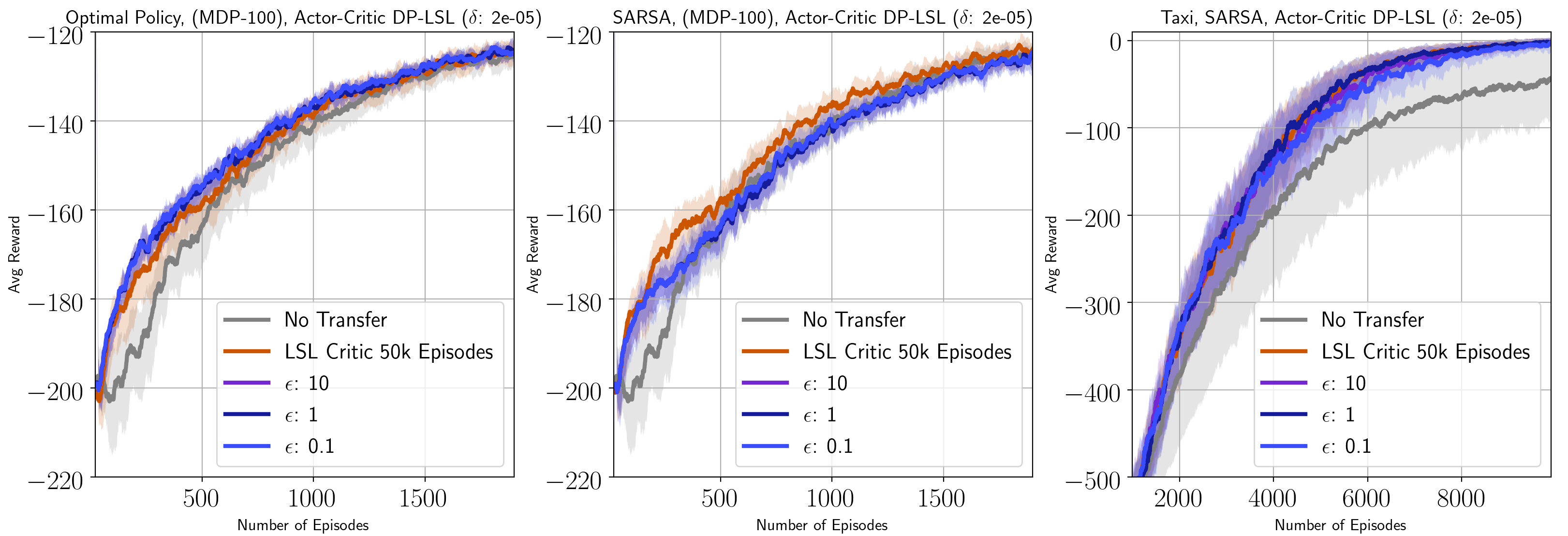}
    \caption{MDP-100 and Taxi-V2 Task Transfer Results under DP-LSL}
    \label{fig:plot_dp_lsl}
\end{figure}

\paragraph{Patient Treatment Progression}
The Markov-Decision Process (MDP) experiments can be regarded as clinical in nature -- patient's data is encoded into a state vector representation similar to \cite{Komorowski2018} -- but it could easily be applied to other domains, such as autonomous navigation. 
We study two approaches to generating samples: taking only optimal actions (the agent selects an action based on argmax $Q^{*}_{\pi}(S,a)$ where the  $Q^{*}_{\pi} $ is the optimal action-value function) and with an on-policy temporal-difference method, SARSA \cite{Sutton2018}. 

Our MDP consists of a chain of $N$ states, where $N=100$. In each state the agent has some probability $p$ of staying and probability $(1-p)$ of advancing (as in supp. material). The environment has two actions: $A_0$ and $A_1$ where their probability of transitioning to the right is $0.1$ and $0.9$ respectively. A reward of 1 is given when the agent reaches the final, absorbing state, and -1 for all other states. States are one-hot encoded in a vector of size 100. This setup illustrates a case of policy evaluation in the medical domain, where patients tend to progress through stages of recovery at different rates, and past states are not usually revisited (because in the medical domain, states contain historic information about past treatments).

\paragraph{Taxi-V2}
\textit{Taxi-V2} \cite{Dietterich2000} is a discrete grid-world environment where the agent must pick up and drop off passengers at the right location. 
%This environment has three stopping conditions: (1) move around the map until the episode ends, incurring -1 reward at every step, (2) pickup a passenger and drop them off at the wrong place with -10 reward and (3) pick up the passenger and drop them off at the right location, receiving 20 reward on termination. 
While this is still a relatively simple environment, it provides a classic scenario where an agent is able to draw on prior experience without leaking information trajectories performed before initialization.

\begin{figure}
    \includegraphics[width=\textwidth]{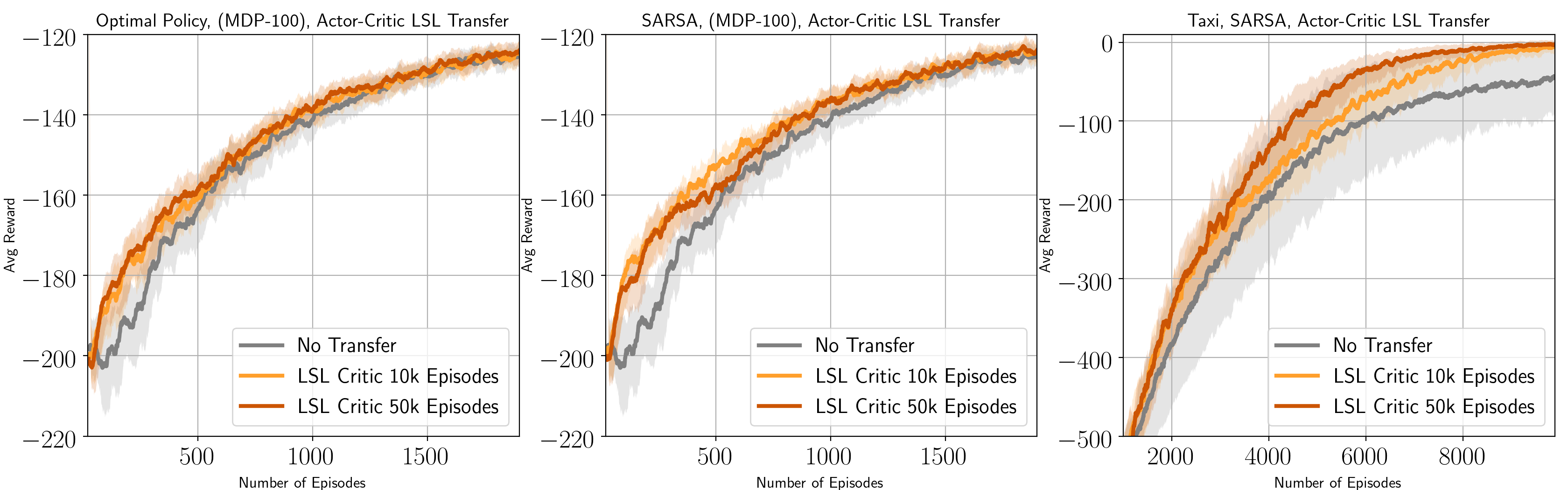}
    \caption{Least Squares Transfer Performance (10k and 50k producer episodes)}
    \label{fig:plot_lsl}
\end{figure}

\paragraph{Results} We find in Figure \ref{fig:plot_dp_lsl} that even with a layer of DP noise, our critic benefits from an initialization of the value function estimate. In an environment where the number of episodes available is finite and regulatory frameworks inhibit sharing data, such a transfer learning approach could provide tangible benefits while minimizing the user harm through parameter sharing. Further experiments in Figure \ref{fig:plot_lsl} illustrate that having more sample episodes $m$ will likely improve the quality of the transfer. We also find that some transfer is better than no transfer. For example in \textit{Taxi-V2}, we achieve convergence after 10,000 episodes, whereas without transfer it took the agent 15,000 episodes. We also note that changing our privacy budget $\epsilon$ by two orders of magnitude does not vary the resulting transfer, meaning that we can benefit from initialization with limited risks to individual privacy. 

\section{Conclusion}

We presented a motivated set of use cases for applying a differentially-private critic in the RL setting. The definition of the producer and consumer trust-model is common in real-world deployments and fits with existing transfer learning approaches where data centralization is difficult. Our preliminary results suggest a measurable improvement in sample efficiency through task transfer. We look forward to exploring how this framework could be extended so that the consumer's critic could then be shared with the producer by leveraging ideas coming from the Federated Learning literature \cite{brendan2018learning}.

\subsubsection*{Acknowledgments}
We thank Joey Bose, Mike Rabbat and Maxime Wabartha for discussion and comments. This work was supported in part by Google DeepMind and Mila.

\bibliographystyle{plain}
\bibliography{references}

\begin{thebibliography}{10}

\bibitem{Abadi2016}
Martin Abadi, Andy Chu, Ian Goodfellow, H.~Brendan McMahan, Ilya Mironov, Kunal
  Talwar, and Li~Zhang.
\newblock {Deep Learning with Differential Privacy}.
\newblock In {\em Proceedings of the 2016 ACM SIGSAC Conference on Computer and
  Communications Security - CCS'16}, pages 308--318, New York, New York, USA,
  2016. ACM Press.

\bibitem{al2019privacy}
Mohammad Al-Rubaie and J~Morris Chang.
\newblock Privacy-preserving machine learning: Threats and solutions.
\newblock {\em IEEE Security \& Privacy}, 17(2):49--58, 2019.

\bibitem{Apple2017}
Apple.
\newblock {Learning with Privacy at Scale}.
\newblock {\em Ml}, 1:1--25, 2017.

\bibitem{Balle2016}
Borja Balle, Maziar Gomrokchi, and Doina Precup.
\newblock {Differentially Private Policy Evaluation}.
\newblock {\em International Conference on Machine Learning (ICML)}, pages
  2130--2138, jun 2016.

\bibitem{bengio2007greedy}
Yoshua Bengio, Pascal Lamblin, Dan Popovici, and Hugo Larochelle.
\newblock Greedy layer-wise training of deep networks.
\newblock In {\em Advances in neural information processing systems}, pages
  153--160, 2007.

\bibitem{openAIGym}
Greg Brockman, Vicki Cheung, Ludwig Pettersson, Jonas Schneider, John Schulman,
  Jie Tang, and Wojciech Zaremba.
\newblock Openai gym, 2016.

\bibitem{Carlini2018}
Nicholas Carlini, Chang Liu, {\'{U}}lfar Erlingsson, Jernej Kos, and Dawn Song.
\newblock {The Secret Sharer: Evaluating and Testing Unintended Memorization in
  Neural Networks}.
\newblock Technical report, Google, 2018.

\bibitem{Dietterich2000}
Thomas~G. Dietterich.
\newblock {Hierarchical Reinforcement Learning with the MAXQ Value Function
  Decomposition}.
\newblock {\em Journal of Artificial Intelligence Research}, 13:227--303, 2000.

\bibitem{Finn2017}
Chelsea Finn, Pieter Abbeel, and Sergey Levine.
\newblock {Model-Agnostic Meta-Learning for Fast Adaptation of Deep Networks}.
\newblock {\em International Conference on Machine Learning (ICML)}, mar 2017.

\bibitem{fox2019reinforcement}
Ian Fox and Jenna Wiens.
\newblock Reinforcement learning for blood glucose control: Challenges and
  opportunities.
\newblock {\em ICML 2019 Workshop RL4RealLife}, 2019.

\bibitem{fridman2018deeptraffic}
Lex Fridman, Benedikt Jenik, and Jack Terwilliger.
\newblock Deeptraffic: Driving fast through dense traffic with deep
  reinforcement learning.
\newblock {\em arXiv preprint arXiv:1801.02805}, 2018.

\bibitem{gajane2017corrupt}
Pratik Gajane, Tanguy Urvoy, and Emilie Kaufmann.
\newblock Corrupt bandits for preserving local privacy.
\newblock {\em arXiv preprint arXiv:1708.05033}, 2017.

\bibitem{Komorowski2018}
Matthieu Komorowski, Leo~A Celi, Omar Badawi, Anthony~C Gordon, and A~Aldo
  Faisal.
\newblock {The Artificial Intelligence Clinician learns optimal treatment
  strategies for sepsis in intensive care}.
\newblock {\em Nature Medicine}, 24(11):1716--1720, 2018.

\bibitem{Lowe2017}
Ryan Lowe, Yi~Wu, Aviv Tamar, Jean Harb, Pieter Abbeel, and Igor Mordatch.
\newblock Multi-agent actor-critic for mixed cooperative-competitive
  environments.
\newblock In {\em Advances in Neural Information Processing Systems 30: Annual
  Conference on Neural Information Processing Systems 2017, 4-9 December 2017,
  Long Beach, CA, {USA}}, pages 6379--6390, 2017.

\bibitem{brendan2018learning}
H.~Brendan McMahan, Daniel Ramage, Kunal Talwar, and Li~Zhang.
\newblock Learning differentially private recurrent language models.
\newblock In {\em International Conference on Learning Representations}, 2018.

\bibitem{Nissim2011}
Kobbi Nissim, Sofya Raskhodnikova, and Adam~D. Smith.
\newblock Smooth sensitivity and sampling in private data analysis.
\newblock In {\em Proceedings of the 39th Annual {ACM} Symposium on Theory of
  Computing, San Diego, California, USA, June 11-13, 2007}, pages 75--84, 2007.

\bibitem{Onu2019}
Charles~C. Onu, Jonathan Lebensold, William~L. Hamilton, and Doina Precup.
\newblock {Neural Transfer Learning for Cry-based Diagnosis of Perinatal
  Asphyxia}.
\newblock {\em Interspeech}, jun 2019.

\bibitem{oquab2014learning}
Maxime Oquab, Leon Bottou, Ivan Laptev, and Josef Sivic.
\newblock Learning and transferring mid-level image representations using
  convolutional neural networks.
\newblock In {\em Proceedings of the IEEE conference on computer vision and
  pattern recognition}, pages 1717--1724, 2014.

\bibitem{Pan2019}
Xinlei Pan, Weiyao Wang, Xiaoshuai Zhang, Bo~Li, Jinfeng Yi, and Dawn Song.
\newblock How you act tells a lot: Privacy-leakage attack on deep reinforcement
  learning.
\newblock {\em CoRR}, abs/1904.11082, 2019.

\bibitem{Parisotto2016}
Emilio Parisotto, Jimmy~Lei Ba, and Ruslan Salakhutdinov.
\newblock {Actor-Mimic: Deep Multitask and Transfer Reinforcement Learning}.
\newblock {\em ICLR}, 2016.

\bibitem{Rusu2015}
Andrei~A Rusu, Sergio~Gomez Colmenarejo, Caglar Gulcehre, Guillaume Desjardins,
  James Kirkpatrick, Razvan Pascanu, Volodymyr Mnih, Koray Kavukcuoglu, and
  Raia Hadsell.
\newblock {Policy Distillation}.
\newblock {\em ICLR 2016}, page~13, 2015.

\bibitem{Shariff2018}
Roshan Shariff and Or~Sheffet.
\newblock Differentially private contextual linear bandits.
\newblock In {\em Advances in Neural Information Processing Systems 31: Annual
  Conference on Neural Information Processing Systems 2018, NeurIPS 2018, 3-8
  December 2018, Montr{\'{e}}al, Canada.}, pages 4301--4311, 2018.

\bibitem{Sutton2018}
Richard~S. Sutton and Andrew~G. Barto.
\newblock {\em Reinforcement Learning: An Introduction}.
\newblock The MIT Press, second edition, 2018.

\bibitem{taylor2009transfer}
Matthew~E Taylor and Peter Stone.
\newblock Transfer learning for reinforcement learning domains: A survey.
\newblock {\em Journal of Machine Learning Research}, 10(Jul):1633--1685, 2009.

\bibitem{Teh2017}
Yee~Whye Teh, Victor Bapst, Wojciech~M. Czarnecki, John Quan, James
  Kirkpatrick, Raia Hadsell, Nicolas Heess, and Razvan Pascanu.
\newblock Distral: Robust multitask reinforcement learning.
\newblock In {\em Advances in Neural Information Processing Systems 30: Annual
  Conference on Neural Information Processing Systems 2017, 4-9 December 2017,
  Long Beach, CA, {USA}}, pages 4496--4506, 2017.

\bibitem{tossou2017achieving}
Aristide Charles~Yedia Tossou and Christos Dimitrakakis.
\newblock Achieving privacy in the adversarial multi-armed bandit.
\newblock In {\em Thirty-First AAAI Conference on Artificial Intelligence},
  2017.

\bibitem{tossou2016algorithms}
Aristide~CY Tossou and Christos Dimitrakakis.
\newblock Algorithms for differentially private multi-armed bandits.
\newblock In {\em Thirtieth AAAI Conference on Artificial Intelligence}, 2016.

\bibitem{Wang2019}
Baoxiang Wang and Nidhi Hegde.
\newblock Private q-learning with functional noise in continuous spaces.
\newblock {\em CoRR}, abs/1901.10634, 2019.

\bibitem{wang2017ACER}
Ziyu Wang, Victor Bapst, Nicolas Heess, Volodymyr Mnih, Remi Munos, Koray
  Kavukcuoglu, and Nando de~Freitas.
\newblock Sample efficient actor-critic with experience replay.
\newblock {\em International Conference on Learning Representations}, 2017.

\bibitem{yosinski2014transferable}
Jason Yosinski, Jeff Clune, Yoshua Bengio, and Hod Lipson.
\newblock How transferable are features in deep neural networks?
\newblock In {\em Advances in neural information processing systems}, pages
  3320--3328, 2014.

\bibitem{zhu2017target}
Yuke Zhu, Roozbeh Mottaghi, Eric Kolve, Joseph~J Lim, Abhinav Gupta,
  Li~Fei-Fei, and Ali Farhadi.
\newblock Target-driven visual navigation in indoor scenes using deep
  reinforcement learning.
\newblock In {\em 2017 IEEE international conference on robotics and automation
  (ICRA)}, pages 3357--3364. IEEE, 2017.

\end{thebibliography}

\end{document}